\documentclass[10pt,twocolumn,letterpaper]{article}

\usepackage{cvpr}
\usepackage{times}
\usepackage{epsfig}
\usepackage{graphicx}
\usepackage{amsmath}
\usepackage{amssymb}

\usepackage{graphicx}
\usepackage{subfigure}
\usepackage{float}
\usepackage{multirow}
\usepackage{overpic}

\usepackage{booktabs}
\usepackage{bm}
\usepackage{threeparttable}

\usepackage{appendix}

\usepackage[breaklinks=true,bookmarks=false]{hyperref}

\cvprfinalcopy 

\def\cvprPaperID{8820} 

\DeclareRobustCommand*{\IEEEauthorrefmark}[1]{%
    \raisebox{0pt}[0pt][0pt]{\textsuperscript{\footnotesize\ensuremath{#1}}}}

\ifcvprfinal\fi
\begin{document}

\title{Real-world Person Re-Identification via Degradation Invariance Learning}

\author{Yukun Huang\IEEEauthorrefmark{1},
Zheng-Jun Zha\IEEEauthorrefmark{1}\thanks{Corresponding author.},
Xueyang Fu\IEEEauthorrefmark{1},
Richang Hong\IEEEauthorrefmark{2},
Liang Li\IEEEauthorrefmark{3}\\
\IEEEauthorrefmark{1}University of Science and Technology of China, China\\
\IEEEauthorrefmark{2}Hefei University of Technology, China\\
\IEEEauthorrefmark{3}Institute of Computing Technology, Chinese Academy of Sciences, China\\
{\tt\small kevinh@mail.ustc.edu.cn, \{zhazj, xyfu\}@ustc.edu.cn, hongrc.hfut@gmail.com, liang.li@ict.ac.cn}
}

\maketitle

\thispagestyle{empty} 

\begin{abstract}
Person re-identification (Re-ID) in real-world scenarios usually suffers from various degradation factors, e.g., low-resolution, weak illumination, blurring and adverse weather. On the one hand, these degradations lead to severe discriminative information loss, which significantly obstructs identity representation learning; on the other hand, the feature mismatch problem caused by low-level visual variations greatly reduces retrieval performance. An intuitive solution to this problem is to utilize low-level image restoration methods to improve the image quality. However, existing restoration methods cannot directly serve to real-world Re-ID due to various limitations, e.g., the requirements of reference samples, domain gap between synthesis and reality, and incompatibility between low-level and high-level methods. In this paper, to solve the above problem, we propose a degradation invariance learning framework for real-world person Re-ID. By introducing a self-supervised disentangled representation learning strategy, our method is able to simultaneously extract identity-related robust features and remove real-world degradations without extra supervision. We use low-resolution images as the main demonstration, and experiments show that our approach is able to achieve state-of-the-art performance on several Re-ID benchmarks. In addition, our framework can be easily extended to other real-world degradation factors, such as weak illumination, with only a few modifications.
\end{abstract}

\section{Introduction}
\begin{figure}[t]
\centering
\begin{minipage}[b]{\textwidth}
\includegraphics[width=0.50\textwidth]{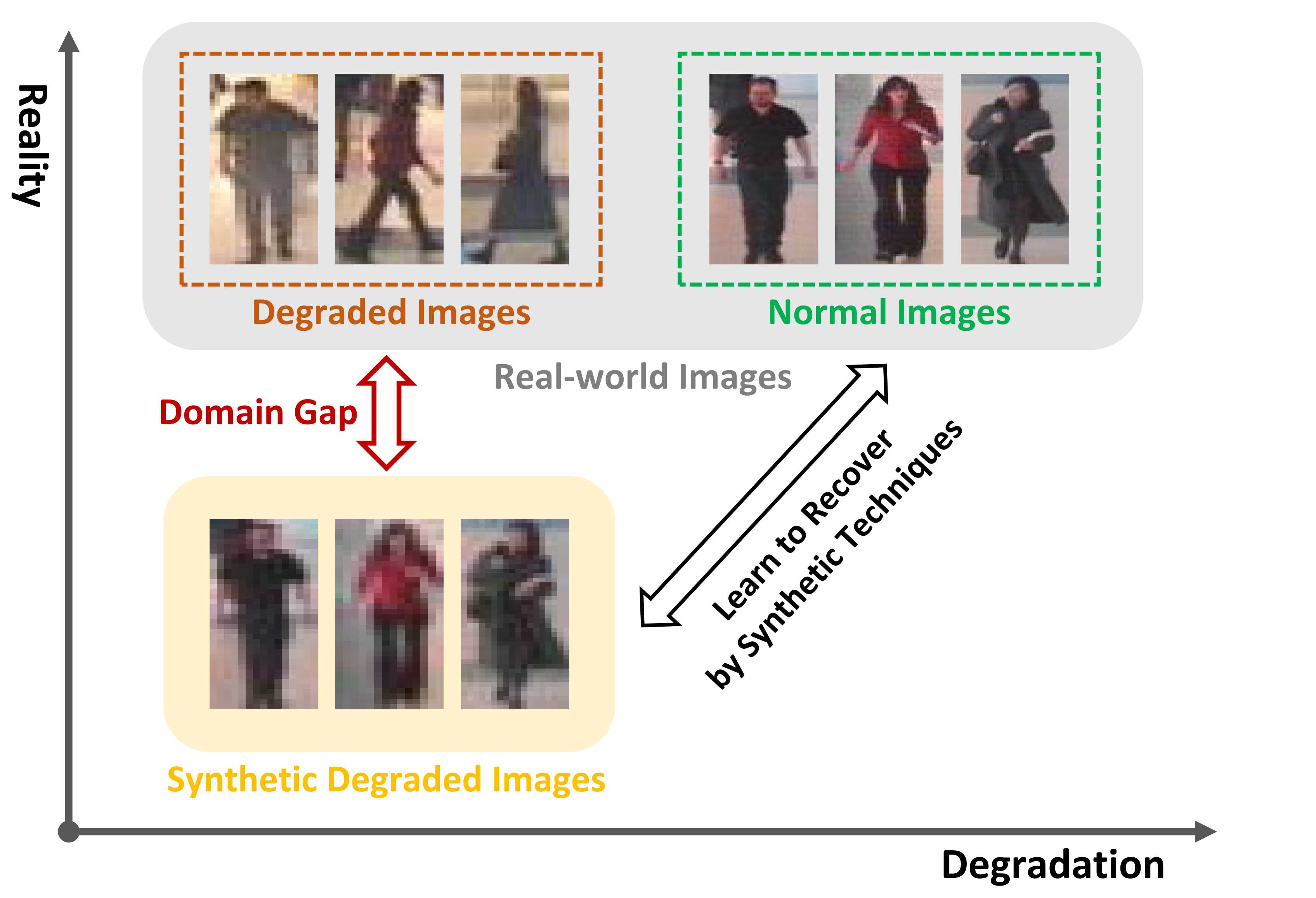}
\end{minipage}
\caption{Existing methods \cite{li2019recover, jiao2018deep, huang2019mm} only use simple synthetic techniques, such as down-sampling for low resolution or gamma correction for low illumination, to alleviate the image degradation issue in Re-ID. This is far from the complex degradation in real-world scenarios, which leads to the domain gap.}
\label{fig1}
\end{figure}

Person re-identification (Re-ID) is a pedestrian retrieval task for non-overlapping camera networks. It is very challenging since the same identity captured by different cameras usually have significant variations in human pose, view, illumination conditions, resolution and so on. To withstand the interference of identity-independent variations, the major target of person Re-ID is to extract robust identity representations. With the powerful representation learning capability, deep convolutional neural networks-based methods have achieved remarkable performance on publicly available benchmarks. For example, the rank-1 accuracy on Market-1501 \cite{zheng2015scalable} has reached 94.8\% \cite{zheng2019joint}, which is very close to the human-level performance.

However, there are still some practical issues that need to be solved for real-world surveillance scenarios, and low quality images caused by various degradation factors are one of them. Several previous works \cite{huang2019mm, mao2019resolution} have demonstrated that such degradations have a serious negative impact on the person Re-ID task. On the one hand, these degradations lead to pool visual appearances and discriminative information loss, making representation learning more difficult; on the other hand, it brings the feature mismatch problem and greatly reduces the retrieval performance.

Existing methods, which focus on alleviating the low-level degradation issue, can be classified into three types:

\textbf{1) Data augmentation.} This kind of methods \cite{bak2018domain} synthesize more training samples under different low-level visual conditions to improve the generalization performance of the model. However, there is a domain gap between synthetic data and real-world data. For example, most of cross-resolution person Re-ID works use the simple down-sampling operator to generate low-resolution images. While the real-world low-resolution images captured usually contain more degradations, such as noise and blurring.

\textbf{2) Combination with low-level vision tasks.} This type of methods \cite{jiao2018deep, wang2018cascaded, mao2019resolution, huang2019mm}, which usually consists of a two-stage pipeline, \ie, combine Re-ID backbone with existing image restoration or enhancement modules to eliminate the effects of degradations. Nevertheless, most existing low-level vision algorithms require aligned training data, which is impossible to collect in real-world surveillance scenarios.

\textbf{3) Disentangled representation learning.} In recent years, some studies attempt to utilize generative adversarial networks (GANs) to learn disentangled representations, which is invariant to some certain interference factors, \eg, human pose \cite{ge2018fd} or resolution \cite{chen2019learning, li2019recover}. Ge \textit{et al.} \cite{ge2018fd} propose FD-GAN for pose-invariant feature learning without additional pose annotations of the training set. To guide the extraction of disentangled features, auxiliary information usually needs to be introduced, which inevitably leads to additional estimation errors or domain bias.

Based on the above observations, we argue that the lack of supervised information about real degradations is the main difficulty in solving real-world Re-ID. This inspires us to think about how to adaptively capture the real-world degradations with limited low-level supervision information. In this work, we propose a Degradation-Invariant representation learning framework for real-world person Re-ID, named \textbf{DI-REID}. With self-supervised and adversarial training strategies, our approach is able to preserve identity-related features and remove degradation-related features. The DI-REID consists of: (a) a content encoder and a degradation encoder to extract \textbf{content} and \textbf{degradation} features from each pedestrian image; (b) a decoder to generate images from previous features; (c) a reality discriminator and a degradation discriminator to provide domain constraints.

To effectively capture the real-world degradations, we generate images by switching the content or degradation features of self-degraded image pairs and real image pairs. The reality discriminator is employed to reduce the domain gap between the synthesis and reality, while the degradation discriminator aims to estimate the degree of degradation of inputs. Utilizing these two discriminators is beneficial for degradation-invariant representation learning. Since the degradation degree may not have a certain discrete division, we use rankGAN \cite{lin2017adversarial} as the degradation discriminator to solve this problem.


In summary, our contribution is three-fold:
\begin{itemize}
\item We introduce a new direction to improve the performance of person re-identification affected by various image degradations in real-world scenarios. Our method can alleviate the need for large amounts of labeled data in existing image restoration methods.
\item We propose a degradation invariance learning framework to extract robust identity representations for real-world person Re-ID. With the self-supervised and disentangled representation learning, our method is able to capture and remove the real-world degradations without extra labeled data.
\item Experiments on several challenging Re-ID benchmarks demonstrate that our approach favorably performs against the state-of-the-art methods. With a few modifications, our method is able to cope with different kinds of degraded images in real-world scenarios.

\end{itemize}

\section{Related Work}
Since our work is related with feature representation learning and GANs, we first briefly summarize these two aspects of works.

\subsection{Feature Representation Learning}
Person re-identification, including image-based Re-ID \cite{li2014deepreid, zheng2015scalable} and video-based Re-ID \cite{zhang2018learning, liu2019dense}, is a very challenging task due to dramatic variations of human pose, camera view, occlusion, illumination, resolution and so on. An important objective of Re-ID is to learn identity representations, which are robust enough for the interference factors mentioned above. These interference factors can be roughly divided into high-level variations and low-level variations.

\textbf{Feature learning against high-level variations.} Such variations include pose, view, occlusions, \etc. Since these variations tend to be spatially sensitive, one typical solution is to leverage local features, \ie, pre-defined regional partition \cite{Miao_2019_ICCV, sun2019perceive, sun2018beyond, liu2018ca3net}, multi-scale feature fusion \cite{liu2016multi, chen2017person, zhou2019omni}, attention-based models \cite{Chen_2019_ICCV, abdnet19iccv, li2018harmonious, liu2017hydraplus, zhao2017deeply} and semantic parts extraction \cite{Guo_2019_ICCV, Miao_2019_ICCV, kalayeh2018human, zhao2017spindle}. These methods usually require auxiliary tasks, such as pose estimation or human parsing. The research line described above has been fully explored and will not be discussed in detail. In this work, we focus on the low-level variation problem.

\textbf{Feature learning against low-level variations.} Such variations include illumination, resolution, weather, \etc. Low-level variations tend to have global consistency and can be alleviated by image restoration methods, such as super-resolution or low-light enhancement.

Most existing Re-ID methods, which are developed for low-level variations, focus on the cross-resolution issue. Jiao  \etal \cite{jiao2018deep} propose to optimize SRCNN and Re-ID network simultaneously in an end-to-end fashion. It is the first work to introduce super-resolution methods to deal with low-resolution Re-ID. To improve the scale adaptability of SR methods, Wang \etal \cite{wang2018cascaded} adopt the cascaded SRGAN structure to progressively recover lost details. Mao \etal \cite{mao2019resolution} propose Foreground-Focus Super-Resolution module to force the SR network to focus on the human foreground, then a dual stream module is used to extract resolution-invariant features. On the other hand, several methods \cite{chen2019learning, li2019recover} utilize adversarial learning to extract the resolution-invariant representations.

Similar to the resolution issue, illumination is another common problem in real-world scenarios. The main impact of illumination variations is the change in color distribution, which has been studied in \cite{varior2016learning, kviatkovsky2012color}. In view of the lack of illumination diversity in the current re-identification datasets, Bak \etal \cite{bak2018domain} introduce SyRI dataset, which provides 100 virtual humans rendered with different illumination maps. Based on the Retinex theory, Huang \etal \cite{huang2019mm} propose a joint framework of Retinex decomposition and person Re-ID to extract illumination-invariant features.

\subsection{Generative Adversarial Networks}

Generative Adversarial Network is first proposed by Goodfellow \textit{et al.} \cite{goodfellow2014generative} to estimate generative models, and then spawn a large number of variants and applications. For person Re-ID, the usage of GANs can be roughly divided into three categories: domain transfer \cite{zhong2018camera, wei2018person, deng2018image, bak2018domain, liu2019adaptive, wang2019learning}, data augmentation \cite{ma2018disentangled, li2018adversarial, qian2018pose, zheng2019joint, hou2019vrstc} and feature representation learning \cite{ge2018fd, liu2019deep, chen2019learning, li2019recover, zha2020adversarial}. Liu \textit{et al.} \cite{liu2019adaptive} utilize multiple GANs to perform factor-wise sub-transfers and achieves superior performance over other unsupervised domain adaptation methods. Zheng \textit{et al.} \cite{zheng2019joint} integrate the discriminative model and the generative model into a unified framework to mutually benefit the two tasks. Hou \textit{et al.} \cite{hou2019vrstc} propose STCnet to explicitly recover the appearance of the occluded areas based on the temporal context information. Similar to STCnet, Li \textit{et al.} \cite{li2019recover} propose Cross-resolution Adversarial Dual Network to simultaneously reconstruct the missing details and extract resolution-invariant features.

\begin{figure*}[t]
  \centering
\begin{minipage}[b]{0.85\textwidth}
\includegraphics[width=\textwidth]{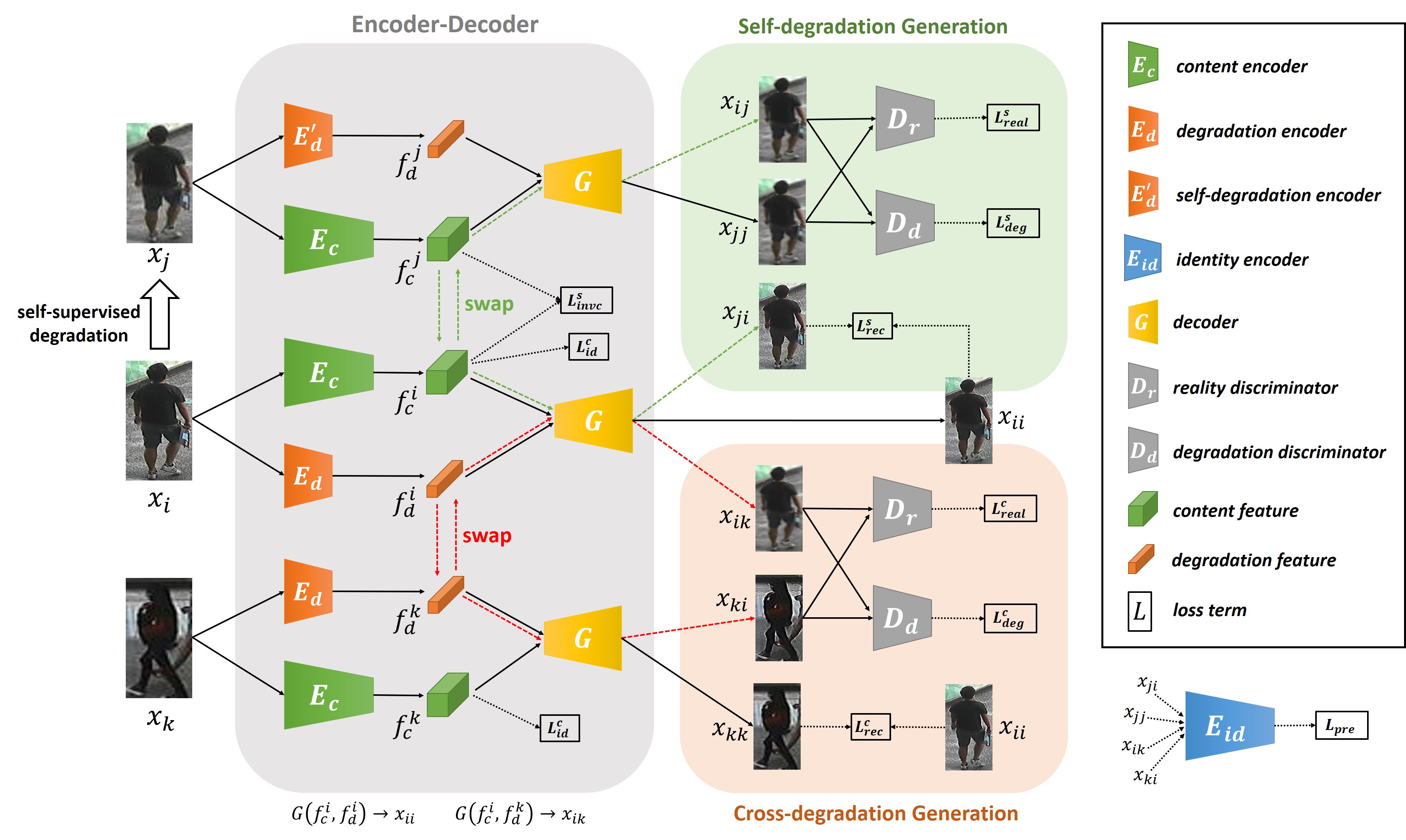}
\end{minipage}
\caption{Overview of the proposed Degradation Decomposition Generative Adversarial Network, DDGAN.
A self-degraded image pair $\{x_i, x_j\}$ and a real image pair $\{x_i, x_k\}$ are alternately used to train the DDGAN. For each pair, input images are decomposed into content features $f_c$ and degradation features $f_d$,
which are then swapped and combined to generate four reconstructed images, \eg, $\{x_{ii}, x_{jj}, x_{ij}, x_{ji}\}$.
}
\label{ddgan}
\end{figure*}

\section{Proposed Method}

\subsection{Overview}
\label{overview}
\begin{figure}[t]
  \centering
\begin{minipage}[b]{0.50\textwidth}
\includegraphics[width=\textwidth]{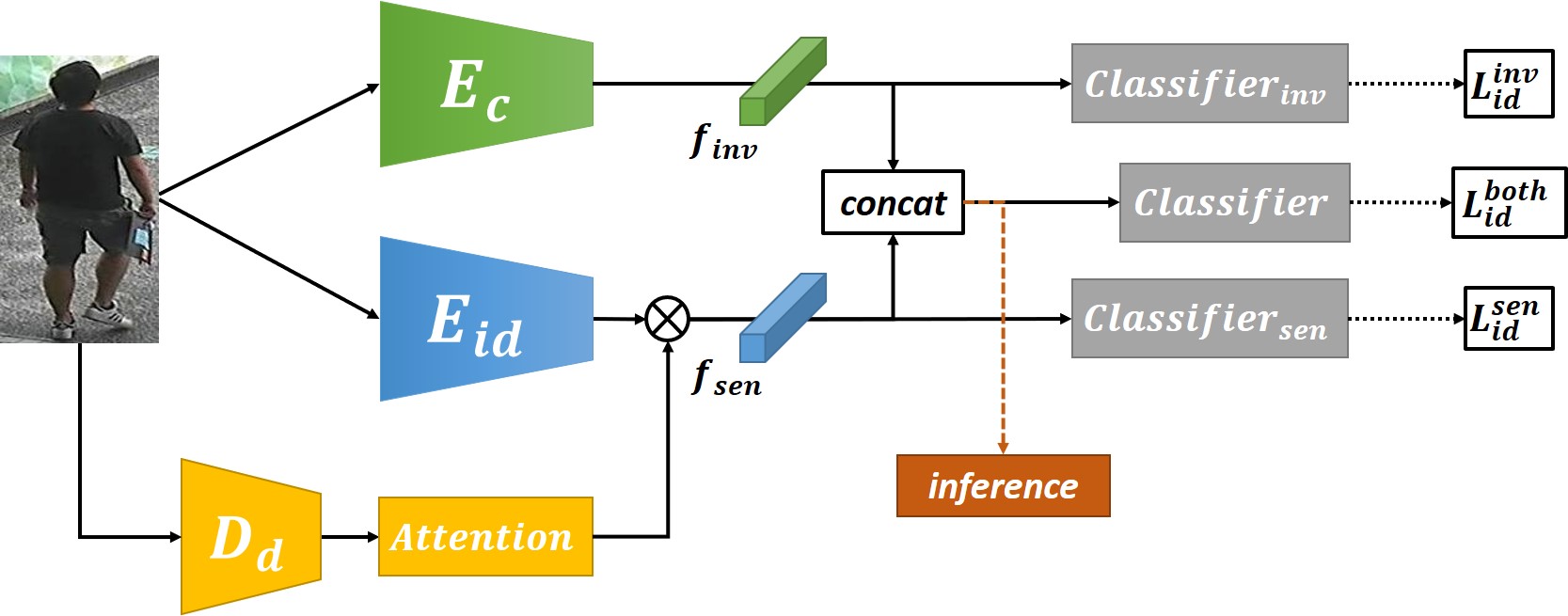} \\
\end{minipage}
\caption{Overview of proposed Dual Feature Extraction Network (DFEN) for robust identity representation learning.}
\label{dfen}
\end{figure}

As shown in Figures \ref{ddgan} and \ref{dfen}, our proposed DI-REID consists of two stages: a \textbf{degradation invariance learning} by a Degradation Decomposition Generative Adversarial Network (DDGAN) and a robust \textbf{identity representation learning} by a Dual Feature Extraction Network (DFEN).

To learn degradation-invariant representations, we attempt to capture and separate the real-world degradation component from a single image. This is an ill-posed problem and extremely difficult since there are no degradation annotations or reference images in the real-world scenarios. Therefore, we synthesize self-degraded images to provide prior knowledge and guidance with self-supervised methods, such as down-sampling, gamma correction and so on. During the degradation invariance learning stage, the aligned self-degraded image pairs and the non-aligned real image pairs are used to train DDGAN in turn, which helps to narrow the domain gap between synthesis and reality.

For identity representation learning, we find that using only degradation-invariant representations does not lead to superior Re-ID performance. This is because degradation invariance forces the network to abandon those discriminative but degradation-sensitive features, \eg, color cues to illumination invariance. Therefore, we design a dual feature extraction network to simultaneously extract both types of features. Besides, an attention mechanism is introduced for degradation-guided feature selection.

\subsection{Network Architecture}

\textbf{Content Encoder} \bm{$E_{c}$}. The content encoder $E_{c}$ is used to extract content features for image generation as well as degradation-invariant identity representation, and DDGAN and DFEN share the same content encoder. In particular, a multi-scale structure is employed for $E_{c}$ to facilitate gradient back propagation.

\textbf{Degradation Encoders} \bm{$E_{d}$} and \bm{$E_{d}'$}. Due to the domain gap between real-world images and self-degraded images, we design a degradation encoder $E_{d}$ and a self-degradation encoder $E_{d}'$  to capture the degradation information, respectively. Note that the weights of $E_{d}$ and $E_{d}'$ are not shared, and $E_{d}'$ is encouraged to convert synthetic degradation features into real-world degradation features.

\textbf{Decoder} \bm{$G$}. Similar to \cite{zheng2019joint}, we utilize the adaptive instance normalization (AdaIN) layers \cite{huang2017adain} to fuse content and degradation features for image generation.

\textbf{Reality discriminator} \bm{$D_{r}$}. The reality discriminator $D_{r}$ forces the decoder to generate images that are close to the realistic distribution. This can indirectly facilitate the self-degradation encoder $E_{d}'$ to produce real-world degradation features.

\textbf{Degradation discriminator} \bm{$D_{d}$}. The degradation discriminator resolves the degree of degradation of the input, encouraging the encoders to learn disentangled content and degradation representations.

\textbf{Identity Encoder} \bm{$E_{id}$}. As a pre-trained Re-ID backbone network, the identity encoder provides identity preserving constraints for degradation invariance learning. This encoder is used to extract discriminative but degradation-sensitive features during the identity representation learning phase.

\subsection{Degradation Invariance Learning}

We aim to propose a general degradation invariance learning network against various real-world degradations under limited supervised information. In this section, we only describe the most common unsupervised DDGAN. More details about the semi-supervised DDGAN for unpaired data are given in the supplement.

\textbf{Formulation.} Our proposed DDGAN is alternately trained by a self-degraded image pair $p^{s}=\{x_{i},x_{j}\}$ and a real image pair $p^{c}=\{x_{i},x_{k}\}$, which are referred to as {Self-degradation Generation} and {Cross-degradation Generation}. For example, as shown in Figure \ref{ddgan}, during the self-degradation generation phase, the input pair $\{x_{i},x_{j}\}$ is decomposed into content features $\{f_{c}^{i},f_{c}^{j}\}$ and degradation features $\{f_{d}^{i},f_{d}^{j}\}$ by the encoders $E_{c}$, $E_{d}$ and $E_{d}'$. After that, all features are combined in pairs to generate new images $\{x_{ii},x_{ij},x_{jj},x_{ji}\}$ by the decoder $G$, where $x_{ij}$ is generated from $G(f_{c}^{i}, f_{d}^{j})$.

\subsubsection{Self-degradation Generation}

Given a self-degraded image pair $p^{s}=\{x_{i},x_{j}\}$, where $x_{j}=F_{deg}(x_i)$, the type of the self-supervised degradation function $F_{deg}$ depends on the specific real-world degradation factors. Since $x_{j}$ and $x_{j}$ are pixel-wise aligned, their content features should be consistent. We provide this constraint using a invariable content loss:
\begin{equation}
  L_{invc}^{s} = || E_{c}(x_i) - E_{c}(x_j) ||_{1}.
  \label{invconloss}
\end{equation}
Further, we can reconstruct the images $x_{ii}$ and $x_{ji}$ with a pixel-wise reconstruction loss:
\begin{equation}
  L_{recon}^{s} = || G(f_{c}^{i},f_{d}^{i}) - x_i ||_{1} + || G(f_{c}^{j},f_{d}^{i}) - x_i ||_{1}.
  \label{invconloss}
\end{equation}
Note that $L_{recon}^{s}$ should not be applied to the reconstructed images $x_{ij}$ and $x_{jj}$ due to the adaptive effects of the self-degradation encoder $E_{d}'$.

To ensure that the appearance of the reconstructed pedestrian images does not change significantly, an identity feature preserving loss is adopted:
\begin{equation}
  \begin{split}
  L_{pre}^{s} = & || E_{id}(G(f_{c}^{i},f_{d}^{j})) - E_{id}(x_{i}) ||_{1}\\
                        & + || E_{id}(G(f_{c}^{j},f_{d}^{i})) - E_{id}(x_{j}) ||_{1}.
  \label{preserveloss}
    \end{split}
\end{equation}

As mentioned earlier, the self-supervised degradation function $F_{deg}$ tends to introduce undesired domain bias between reality and synthesis, which leads to the learned features to deviate from the real-world distribution. To alleviate this issue, we introduce a reality adversarial loss:
\begin{equation}
\small
  \begin{split}
  L_{real}^{s} = & \mathbb{E}[    log(D_{r}(x_{i}))    +    log(1-D_{r}(G(f_{c}^{i},f_{d}^{j})))    ] \\
   & + \mathbb{E}[                          log(D_{r}(x_{k}))   +    log(1-D_{r}(G(f_{c}^{j},f_{d}^{j})))    ],
  \label{realadvloss}
  \end{split}
\end{equation}
where both $x_{i}$ and $x_{k}$ are real-world images.

At last, our main objective is to learn a degradation independent representation. In other words, after switching the content features of the input image pair, the degradation score ranking of reconstructed images should be consistent with the original ranking. To provide such a constraint, we introduce a degradation ranking loss:
\begin{equation}
\small
  \begin{split}
  L_{deg}^{s} = & max(0,(D_{d}(x_i)-D_{d}(G(f_{c}^{i},f_{d}^{j})))*\gamma+\epsilon) \\
   		       & + max(0,(D_{d}(G(f_{c}^{j},f_{d}^{i}))-D_{d}(x_j))*\gamma+\epsilon),
  \label{degadvloss}
  \end{split}
\end{equation}
where $\gamma = 1$ is the rank label of the input image pair, and the margin $\epsilon$ controls the difference of degradation scores. A higher degradation score means lower image quality.

\subsubsection{Cross-degradation Generation}

For the cross-degradation generation, we also perform image encoding and decoding on the input real image pair $p^{c}=\{x_{i},x_{k}\}$, where $x_{i}$ and $x_k$ are directly sampled from the real-world data. To provide the regularization constraint, we also introduce a self-reconstruction loss:
\begin{equation}
  L_{recon}^{c} = || G(f_{c}^{i},f_{d}^{i}) - x_i ||_{1} + || G(f_{c}^{k},f_{d}^{k}) - x_k ||_{1},
  \label{invconloss}
\end{equation}
a reality adversarial loss:
\begin{equation}
\small
  \begin{split}
  L_{real}^{c} = & \mathbb{E}[    log(D_{r}(x_{i}))    +    log(1-D_{r}(G(f_{c}^{i},f_{d}^{k})))    ] \\
   &+ \mathbb{E}[                          log(D_{r}(x_{k}))   +    log(1-D_{r}(G(f_{c}^{k},f_{d}^{i})))    ],
  \label{realadvloss}
  \end{split}
\end{equation}
and an identity feature preserving loss:
\begin{equation}
\small
  \begin{split}
  L_{pre}^{c} = & || E_{id}(G(f_{c}^{i},f_{d}^{k})) - E_{id}(x_{i}) ||_{1} \\
                        & + || E_{id}(G(f_{c}^{k},f_{d}^{i})) - E_{id}(x_{k}) ||_{1}.
  \label{preserveloss}
    \end{split}
\end{equation}
Different from self-degradation generation, $x_i$ and $x_k$ here have completely inconsistent content information, which means the invariable content loss is no longer applicable.

Since the purpose of degradation invariance learning is to improve the real-world person Re-ID, we use a standard identification loss to provide task-driven constraints:
\begin{equation}
  \begin{split}
  L_{id}^{c} &= \mathbb{E}[-log(p(y_{i}|x_{i}))]+\mathbb{E}[-log(p(y_{k}|x_{k}))],
  \label{degadvloss}
  \end{split}
\end{equation}
where the predicted probability $p(y_{i}|x_{i})$ and $p(y_{k}|x_{k})$ are based on the content features $f_{c}^{i}$ and $f_{c}^{k}$, respectively.

For unsupervised degradation invariance learning, the real-world training data does not have any degradation-related supervised information. To take advantages of real data to model the real-world degradation distribution, we also introduce a degradation ranking loss:
\begin{equation}
  \begin{split}
  L_{deg}^{c} =& max(0,(D_{d}(x_i)-D_{d}(G(f_{c}^{i},f_{d}^{k})))*\gamma+\epsilon) \\
   		       & + max(0,(D_{d}(G(f_{c}^{k},f_{d}^{i}))-D_{d}(x_k))*\gamma+\epsilon),\\
   		      &
\left\{
             \begin{array}{l}
             \gamma=-1 \ \ \          if \ \ \ D_{d}(x_i) > D_{d}(x_k)                                                    \\
             \gamma=1 \ \ \ \ \ \   if \ \ \ D_{d}(x_i) < D_{d}(x_k)
             \end{array}
\right.,
  \end{split}
\end{equation}
where the rank label $\gamma$ depends on the predicted degradation scores of the real-world images $x_i$ and $x_k$. In this way, the disentangled content and degradation features can be learned to approximate the real-world distribution without extra supervised information.

\subsubsection{Optimization}
For self-degradation generation, the total objective is:
\begin{equation}
  \begin{split}
  L_{total}^{s} = & \lambda_{invc} L_{invc}^{s} + \lambda_{recon} L_{recon}^{s} + \lambda_{pre} L_{pre}^{s}  \\
                           & + \lambda_{real} L_{real}^{s} + \lambda_{deg} L_{deg}^{s}.
  \end{split}
\end{equation}
For cross-degradation generation, the total objective is:
\begin{equation}
  \begin{split}
  L_{total}^{c} = & \lambda_{id} L_{id}^{c} + \lambda_{recon} L_{recon}^{c} + \lambda_{pre} L_{pre}^{c}  \\
                           & + \lambda_{real} L_{real}^{c} + \lambda_{deg} L_{deg}^{c}.
  \end{split}
\end{equation}
These two optimization phases are performed alternately.

\begin{table*}[t]
  \caption{Cross-resolution Re-ID performance (\%) compared to the state-of-the-art methods on the MLR-CUHK03, MLR-VIPeR and CAVIAR datasets, respectively.}
  \label{tab_res}
  \begin{center}
    \begin{threeparttable}
  \begin{tabular}{ l | ccc | ccc | ccc }
    \toprule
    \multirow{2}{*}{Method}  				& \multicolumn{3}{c|}{MLR-CUHK03}      & \multicolumn{3}{c|}{MLR-VIPeR}      &  \multicolumn{3}{c}{CAVIAR}           \\
    ~                                                				& Rank-1 & Rank-5 & Rank-10                    & Rank-1 & Rank-5 & Rank-10                & Rank-1 & Rank-5 & Rank-10              \\
    \midrule
    CamStyle \cite{zhong2018camera}		& 69.1 & 89.6 & 93.9                                     & 34.4 & 56.8 & 66.6                                    & 32.1 & 72.3 & 85.9                                \\
    FD-GAN \cite{ge2018fd}                  		& 73.4 & 93.8 & 97.9                                     & 39.1 & 62.1 & 72.5                                    & 33.5 & 71.4 & 86.5                                \\
    \midrule
    JUDEA \cite{li2015multi}				& 26.2 & 58.0 & 73.4                                     & 26.0 & 55.1 & 69.2                                    & 22.0 & 60.1 & 80.8                               \\
    SLD$^{2}$L \cite{jing2015super}		& -        & -        & -                                            & 20.3 & 44.0 & 62.0                                    & 18.4 & 44.8 & 61.2                                \\
    SDF \cite{wang2016scale}				& 22.2 & 48.0 & 64.0                                     & 9.3   & 38.1 & 52.4                                    & 14.3 & 37.5 & 62.5                                \\
    SING \cite{jiao2018deep} 				& 67.7 & 90.7 & 94.7                                     & 33.5 & 57.0 & 66.5                                    & 33.5 & 72.7 & 89.0                                \\
    CSR-GAN \cite{wang2018cascaded}          & 71.3 & 92.1 & 97.4                                     & 37.2 & 62.3 & 71.6                                    & 34.7 & 72.5 & 87.4                                \\
    FFSR+RIFE \cite{mao2019resolution}       & 73.3 & 92.6 & -                                            & 41.6 & 64.9 & -                                           & 36.4 & 72.0 & -                                       \\
    RAIN \cite{chen2019learning}                     & 78.9 & 97.3 & 98.7                                     & 42.5 & 68.3 & 79.6				  & 42.0 & 77.3 & 89.6                                \\
    CAD-Net \cite{li2019recover}                      & 82.1 & \textbf{97.4} & \textbf{98.8}   & 43.1 & 68.2 & 77.5  				  & 42.8 & 76.2 & 91.5                                \\
    \midrule
    ResNet50							& 60.2 & 86.6 & 93.2                                    & 28.5 & 53.8 & 65.2                                   			& 20.2 & 61.0 & 79.8                                   			 	\\
    ResNet50 (tricks\tnote{$\dagger$}\ \ )	& 75.1 & 91.3 & 95.7                                    & 42.1 & 63.9 & 71.5                                   			& 40.6 & 76.2 & 91.0     	                                  			\\
    Ours                                         				& \textbf{85.7} & 97.1 & 98.6                   & \textbf{50.3} & \textbf{77.9} & \textbf{87.3}      	& \textbf{51.2} & \textbf{83.6} & \textbf{94.4}		\\
  \bottomrule
  \end{tabular}
    \begin{tablenotes}
  \item[$\dagger$] Here all tricks we used include RandomHorizontalFlip, RandomCrop, BNNeck \cite{Luo_2019_CVPR_Workshops} and triplet loss.
  \end{tablenotes}
    \end{threeparttable}
  \end{center}
\end{table*}

\subsection{Identity Representation Learning}
As described in \ref{overview}, the DFEN extracts the degradation-invariant features $f_{inv}$ and the degradation-sensitive features $f_{sen}$ as identity representations, where the degradation-invariant features are the content features without dimension reduction.

Given a normal image, both the $f_{inv}$ and $f_{sen}$ should be kept; while for a degraded image, it should keep $f_{inv}$ and suppress $f_{sen}$ for Re-ID. To achieve this goal, we introduce a degradation-guided attention module, which inputs the degradation cues and outputs the attentive weights of $f_{sen}$. Although both $E_{d}$ and $D_{d}$ can provide the degradation information, we choose $D_{d}$ for better interpretability. Given an input image $x_i$, the final identity representation is formulated as:
\begin{equation}
  f_{id}^i = concat(f_{inv}^i,  f_{sen}^i \odot Att(D_{d}(x_i))),
  \label{invconloss}
\end{equation}
where $\odot$ denotes element-wise product.

In addition, we use multiple classifiers to better coordinate these two types of features. The total objective is:
\begin{equation}
  L_{total}^{id} = \lambda_{inv} L_{inv}^{id} + \lambda_{sen} L_{sen}^{id} + \lambda_{both} L_{both}^{id},
  \label{invconloss}
\end{equation}
where each loss term consists of a cross-entropy loss and a triplet loss with hard sample mining \cite{hermans2017defense}.

\section{Experiments}

To evaluate our approach on person Re-ID task against various real-world degradations, we focus on two major degradation factors, \ie, resolution and illumination.

\subsection{Datasets}

We conduct experiments on four benchmarks: CAVIAR \cite{cheng2011custom}, MLR-CUHK03 and MLR-VIPER for {cross-resolution Re-ID}, and MSMT17 \cite{wei2018person} for {cross-illumination Re-ID}.

The \textbf{CAVIAR} dataset comprises 1,220 images of 72 identities captured by two different cameras in an indoor shopping center in Lisbon. Due to the resolution of one camera is much lower than that of the other, it is very suitable for evaluating genuine cross-resolution person Re-ID.

The \textbf{MLR-CUHK03} and \textbf{MLR-VIPeR} datasets are based on the CUHK03 \cite{li2014deepreid} and VIPeR \cite{gray2008viewpoint}, respectively. MLR-CUHK03 includes 14,097 images of 1,467 identities, while MLR-VIPeR contains 632 person image pairs captured from two camera views. Following SING \cite{jiao2018deep}, each image from one camera is down-sampled with a ratio randomly picked from $\{ \frac{1}{2}, \frac{1}{3}, \frac{1}{4} \}$ to construct cross-resolution settings, where the query set consists of LR images while the gallery set is only composed of HR images.

The \textbf{MSMT17} dataset, which contains 32,621/93,820 bounding boxes for training/testing, is collected by 15 surveillance cameras on the campus, including both outdoor and indoor scenes. To cover as many time periods as possible, four days with different weather conditions in one month were selected for collecting the raw video.

\subsection{Implementation Details}
The proposed approach is implemented in PyTorch with two NVIDIA 1080Ti GPUs. All the used images are resized to $256 \times 128 \times 3$. We employ a multi-scale ResNet50 \cite{he2016deep} structure for the content encoder, and both discriminators $D_{real}$ and $D_{deg}$ follow the popular multi-scale PatchGAN structure \cite{isola2017image}.
More details about the optimizations and structures can be found in the supplement.

\subsection{Experimental Settings and Evaluation Metrics}

We employ the single-shot Re-ID settings and use the average Cumulative Match Characteristic (CMC) \cite{jiao2018deep, mao2019resolution, chen2019learning, li2019recover} for evaluating {cross-resolution Re-ID}. In addition, we choose downsampling with a ratio that obeys uniform distribution $U[2, 4]$ as the self-degradation function.

For {cross-illumination Re-ID}, we follow the standard protocols of corresponding datasets. The mean Average Precision (mAP) and CMC are adopted to evaluate the retrieval performance. Gamma correction is used as the self-degradation function, where the gamma value obeys a uniform distribution $U[2, 3.5]$.

\subsection{Re-ID Evaluation and Comparisons}

\textbf{Cross-Resolution.} We compare our DI-REID with the state-of-the-art cross-resolution Re-ID methods as well as standard Re-ID methods. As shown in Table \ref{tab_res}, our approach achieves superior performance on all three adopted datasets and consistently outperform all competing methods at rank-1. Note that our approach outperforms the best competitor \cite{li2019recover} by 8.4\% at rank-1 on the only real-world cross-resolution dataset CAVIAR. It proves the effectiveness of our approach to the real-world resolution degradation.

\textbf{Cross-Illumination.} To demonstrate that our DI-REID is capable of dealing with various real-world degradation, extended evaluation on the real-world MSMT17 dataset is also performed for cross-illumination Re-ID. As reported in Table \ref{tab_ill}, competitive Re-ID performance is also achieved by our approach compared with existing state-of-the-art methods. It is worth mentioning that we only use the illumination degradation prior without introducing extra structural or semantic priors of human body parts.

  \begin{table}
  \begin{center}
    \caption{Cross-illumination Re-ID performance (\%) compared to the state-of-the-art methods on the MSMT17 dataset.}
    \label{tab_ill}
  \begin{tabular}{ l | c|c|c|c }
    \toprule
    Methods  				& Rank-1 & Rank-5 & Rank-10 &mAP       \\
    \midrule
    GoogLeNet \cite{szegedy2015going}        & 47.6 & 65.0 & 71.8 & 23.0 \\
    PDC \cite{su2017pose}                		 & 58.0 & 73.6 & 79.4 & 29.7 \\
    GLAD \cite{wei2017glad}               	 	 & 61.4 & 76.8 & 81.6 & 85.9 \\
    PCB \cite{sun2018beyond}			 & 68.2 & 81.2 & 85.5 & 40.4 \\	
    IANet \cite{hou2019interaction}               	 & 75.5 & 85.5 & 88.7 & 46.8 \\
    \midrule
    ResNet50							& 57.4 & 72.9 & 78.4 & 29.2 \\
    ResNet50 (tricks)					& 68.8 & 80.9 & 84.7 & 35.8 \\
    Ours                                         				& \textbf{75.5} & \textbf{86.2} & \textbf{89.5} & \textbf{47.1}          \\
  \bottomrule
  \end{tabular}
  \end{center}
\end{table}

  \begin{table}
  \begin{center}
  \begin{threeparttable}
    \caption{Ablation Study on the CAVIAR dataset.}
  \label{tab_ablation}
  \begin{tabular}{ l | c|c|c }
  \toprule
    Methods  				& Rank-1 & Rank-5 & Rank-10       \\
  \midrule
   	Ours w/o DIL\footnotemark[1]            	& 44.6 & 82.2 & 93.8                           \\
   	Ours w/o multi-scale            	& 47.2 & 82.4 & 95.2                           \\
 	Ours w/o attention            	& 48.0 & 82.6 & 93.8                           \\
  \midrule
	Ours ($f_{inv}$ only)       	& 41.0 & 80.8 & 92.2                           \\
         Ours ($f_{sen}$ only)      	& 45.4 & 80.0 & 91.2                           \\
         Ours 	            			& 51.2 & 83.6 & 94.4                           \\
  \bottomrule
  \end{tabular}
  \end{threeparttable}
  \end{center}
\end{table}
\footnotetext[1]{For the w/o DIL configuration, we skip the stage of Degradation Invariance Learning (directly assigning ImageNet pretrained weights to the content encoder), and the degradation-guided attention module is disabled.}

\subsection{Feature Analysis and Visualizations}

\begin{figure}[t]
  \centering
\begin{minipage}[b]{0.5\textwidth}
\includegraphics[width=\textwidth]{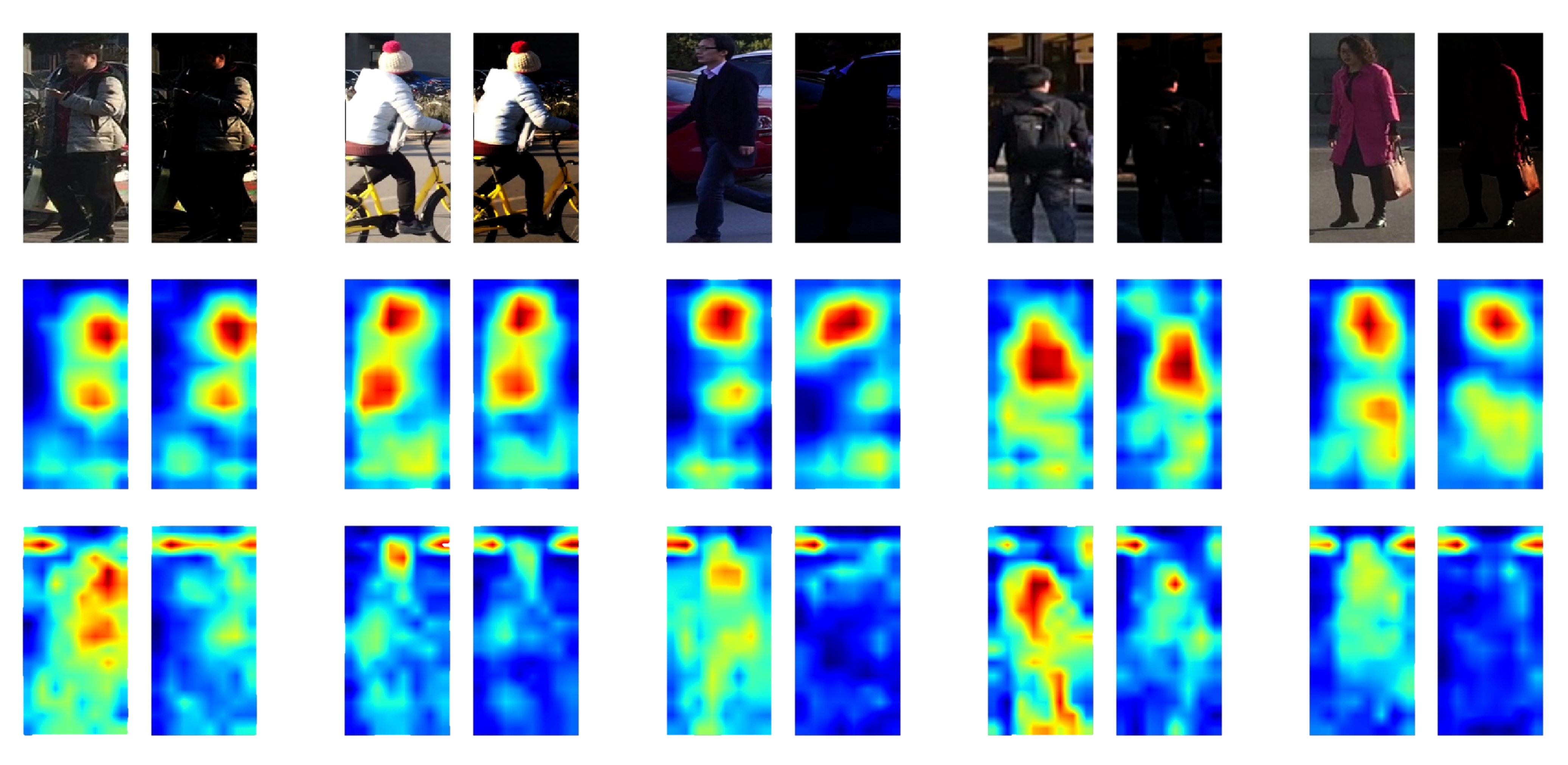}
\end{minipage}
\caption{Visualizations of degradation-invariant features. \textbf{Top}: input images, \textbf{middle}: features produced by our DI-REID, \textbf{bottom}: features produced by a ResNet-50 baseline.}
\label{content}
\end{figure}

\begin{figure}[t]
\centering
\subfigure[Low-Resolution (LR) to High-Resolution (HR).]{\label{inter_res1}\includegraphics[width=0.90\linewidth]{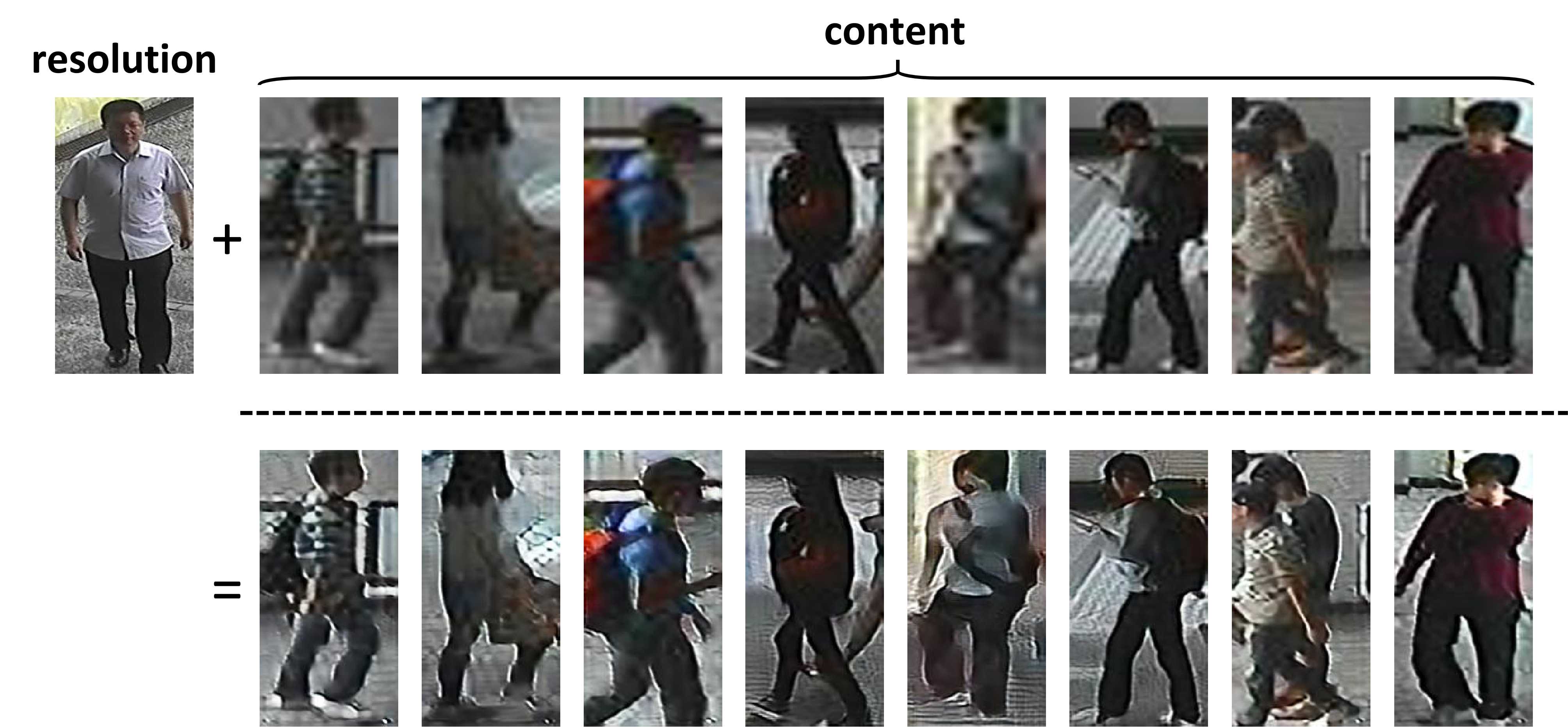}}
\subfigure[High-Resolution to Low-Resolution.]{\label{inter_res2}\includegraphics[width=0.90\linewidth]{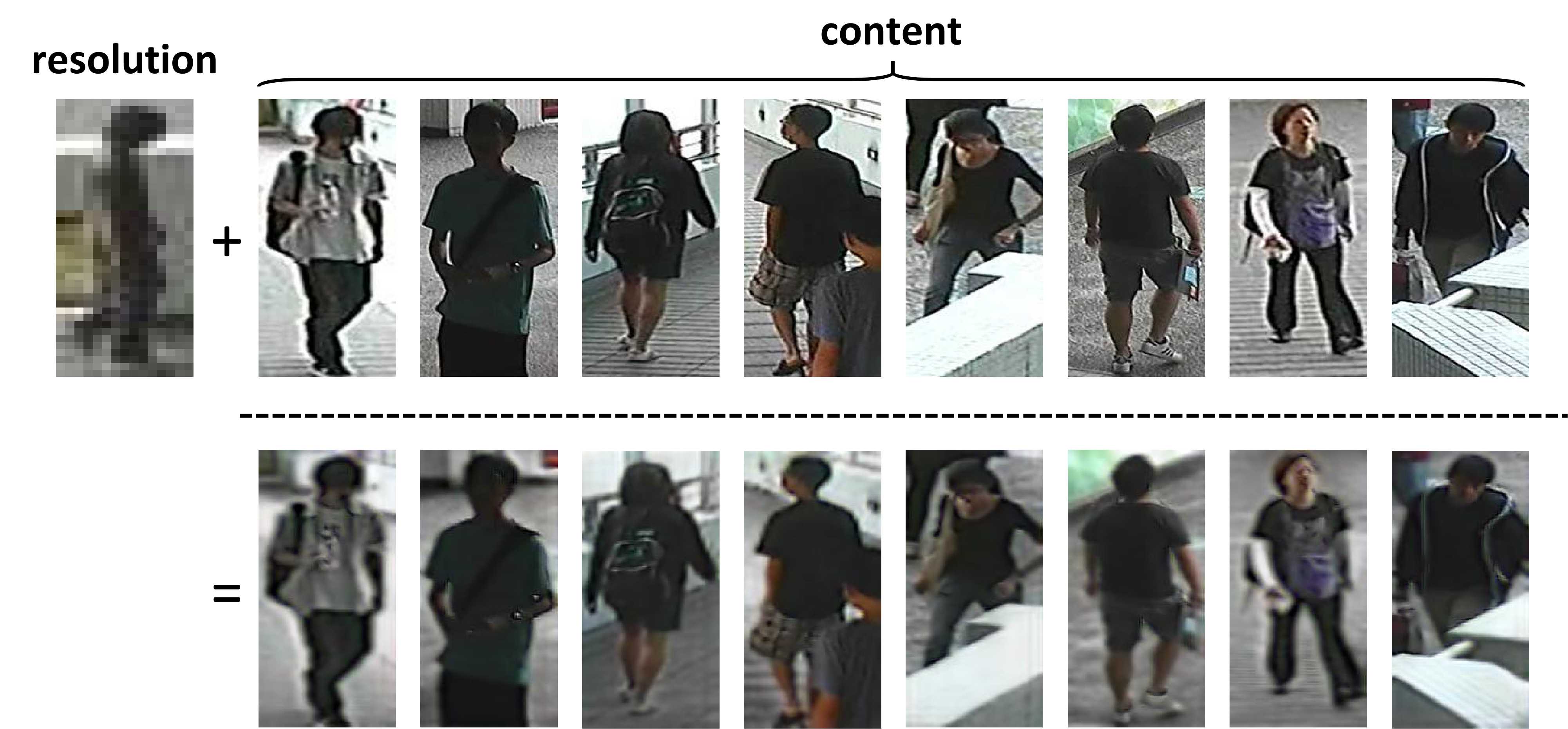}}
\caption{Examples of cross-resolution image generation. (a) Images generated by HR degradation + LR content; (b) Images generated by LR degradation + HR content.}
\label{inter_res}
\end{figure}

\textbf{Degradation-invariant identity features.} We provide the comparison on learned degradation-invariant identity features with a ResNet-50 baseline model. The features are generated by the content encoder and visualizations are shown in Figure \ref{content}. All the feature maps are produced after three downsampling layers for a balance between high-level semantics and fine-grained details. It is clear that despite the degradation of illumination, the attentive regions of degradation-invariant features are basically consistent. Besides, even in extremely low-light conditions (\eg, 6th and 10th columns in Figure \ref{content}), our method still can extract effective discriminative features.
We also find that the learned degradation-invariant features are more focused on local areas,
although no such guidance and constraints are used.

\begin{figure}
  \centering
\begin{minipage}[b]{\textwidth}
\includegraphics[width=0.48\textwidth]{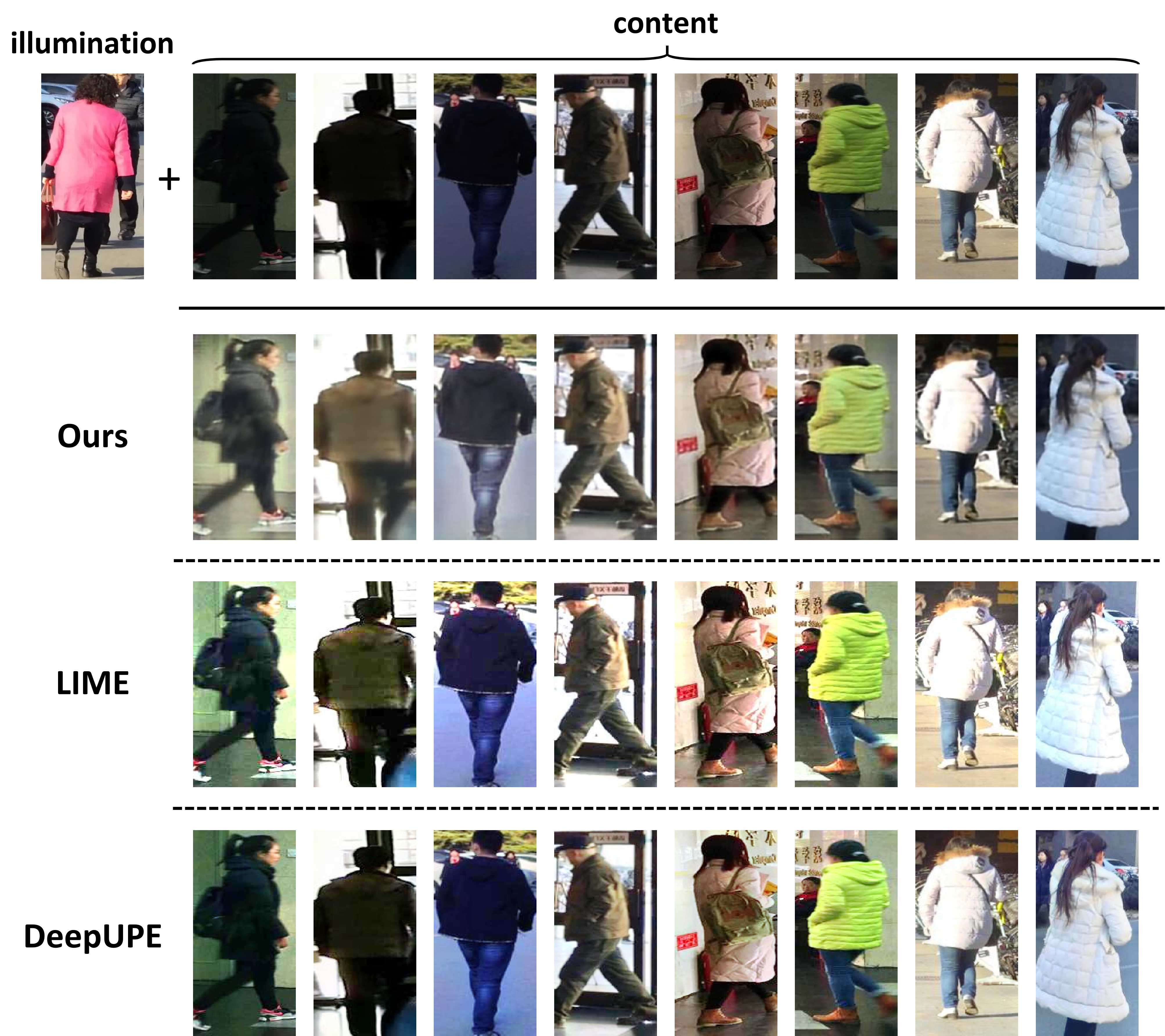} \\
\end{minipage}
\caption{Examples of cross-illumination image generation. The generated results are compared to two state-of-the-art low-light image enhancement methods: LIME \cite{guo2016lime} and DeepUPE \cite{wang2019underexposed}. }
\label{inter_ill}
\end{figure}

\textbf{Analysis of representation disentanglement.}
Since the proposed DI-REID framework extracts degradation-invariant features via disentangled representation learning, it is necessary to analyze the disentangled representations for more insights and interpretability.

As shown in Figures \ref{inter_res} and \ref{inter_ill}, we provide the cross-degradation generation results under cross-resolution and  cross-illumination settings, respectively. By reorganizing content and degradation features, our DDGAN is able to generate new samples with degradation characteristics of the degradation provider and content characteristics of the content provider. In other words, our framework is capable of extracting degradation-independent features as identity representations for the person Re-ID task. In addition, although high-quality image generation is not our purpose, these additional generated samples are expected to be utilized for data augmentation for further performance improvement.

\begin{figure}
  \centering
\begin{minipage}[b]{0.40\textwidth}
\includegraphics[width=\textwidth]{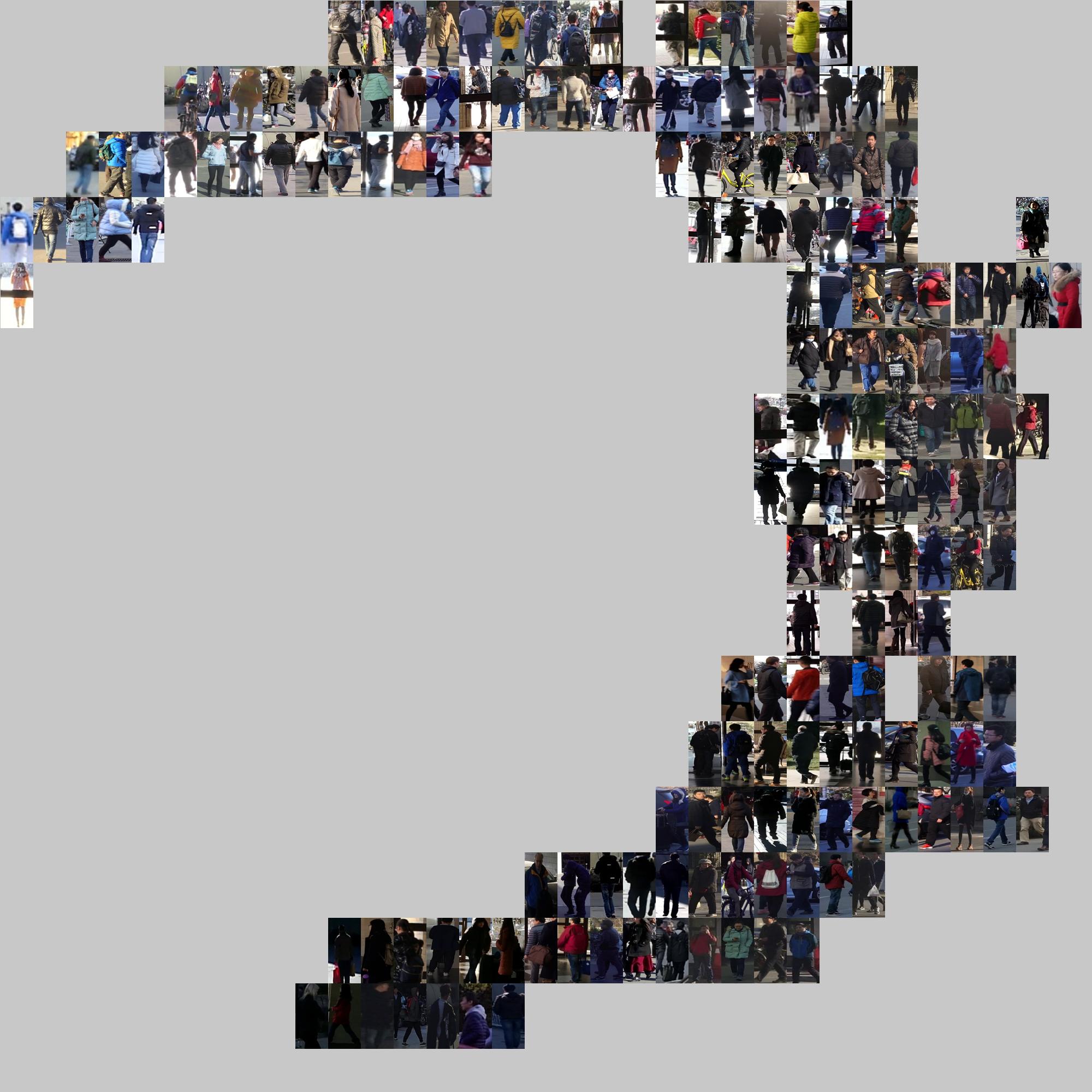}
\end{minipage}
\caption{t-SNE visualization of degradation features on a 1000-sample split, which randomly selected from the MSMT17 dataset.}
\label{tsne}
\end{figure}

\textbf{Analysis of Authenticity.}
Since this work focuses on the real-world degradations, we also analyze the authenticity of degradation features, that is, whether the real-world degradation information is captured. As illustrated in Figure \ref{inter_ill}, our approach achieves very consistent illumination adjustments without causing over-enhancement or under-enhancement. Compared to the existing state-of-the-art low-light image enhancement methods: LIME \cite{guo2016lime} and DeepUPE \cite{wang2019underexposed}, our results are more natural and close to the real-world illumination distribution. We emphasize that our approach does not utilize any of the illumination supervised information of the original dataset, but only with the self-supervised guidance of gamma correction. The t-SNE visualization of learned degradation features of the real-world MSMT17 dataset are shown in Figure \ref{tsne}, and a significant illumination distribution along the manifold can be observed.

\subsection{Ablation Study}
We study the contribution of each component of our approach on the CAVIAR dataset. As shown in Table \ref{tab_ablation}, all the components consistently achieve performance improvements, where the contribution of degradation invariance learning is most significant, resulting in a performance rise of 6.6\% at Rank-1.
We believe the reason is that the learned features is able to simultaneously take account of both identity discriminability and degradation invariance.

We also provide the analysis of degradation-invariant features and degradation-sensitive features, \ie, $f_{inv}$ and $f_{sen}$. It can be observed that $f_{sen}$ performs better at Rank-1, while $f_{inv}$ performs better at Rank-5 and Rank-10.

\section{Conclusion}

In this paper, we propose a degradation-invariance feature learning framework for real-world person Re-ID.
With the capability of disentangled representation and the self-supervised learning, our method is able to capture and remove real-world degradation factors without extra labeled data.
In future work, we consider integrating other semi-supervised feature representation methods, \eg, graph embedding \cite{zhang2014robust}, to better extract pedestrian features from noisy real-world data.


\section{Acknowledgement}


This work was supported by the National Key R\&D Program of China under Grant 2017YFB1300201 and 2017YFB1002203, the National Natural Science Foundation of China (NSFC) under Grants 61622211, U19B2038, 61901433, 61620106009 and 61732007 as well as the Fundamental Research Funds for the Central Universities under Grant WK2100100030.

{\small
\bibliographystyle{ieee_fullname}
\bibliography{egbib}
}

\end{document}